\documentclass[wcp]{jmlr}
\usepackage{amsmath,amssymb,url}
\jmlrvolume{230}
\jmlryear{2024}
\jmlrworkshop{Conformal and Probabilistic Prediction with Applications}
\jmlrproceedings{PMLR}{Proceedings of Machine Learning Research}

\usepackage{aligned-overset}
\usepackage{algorithm}%
\usepackage{xcolor}
\usepackage{algorithm2e}
\usepackage{bbm}
\usepackage{adjustbox}
\usepackage{multirow}
\usepackage{booktabs}
\usepackage{graphicx}

\DeclareMathOperator*{\argmax}{arg\,max}
\DeclareMathOperator*{\argmin}{arg\,min}

\newcommand{\ntrain}{{n_{\textrm{train}}}}

\newcommand{\score}{conformity score}

\title{Entropy Reweighted Conformal Classification}

\author{\Name{Rui Luo}\Email{ruiluo@cityu.edu.hk} \\
       \addr{City University of Hong Kong, Hong Kong, China} \\
       \Name{Nicolo Colombo}\Email{nicolo.colombo@rhul.ac.uk} \\
       \addr{Royal Holloway, University of London, Egham, Surrey, UK}}

\editor{Simone Vantini, Matteo Fontana, Aldo Solari, Henrik Boström and Lars Carlsson}

\begin{document}

\maketitle

\begin{abstract}%
Conformal Prediction (CP) is a powerful framework for constructing prediction sets with guaranteed coverage. However, recent studies have shown that integrating confidence calibration with CP can lead to a degradation in efficiency. In this paper, We propose an adaptive approach that considers the classifier's uncertainty and employs entropy-based reweighting to enhance the efficiency of prediction sets for conformal classification. Our experimental results demonstrate that this method significantly improves efficiency. 
\end{abstract}


\section{Introduction}
Conformal Prediction (CP) is a well-established framework for constructing prediction sets with guaranteed coverage, regardless of the underlying distribution of the data. However, the efficiency of CP prediction sets can be affected by the accuracy of the underlying classifier's uncertainty quantification.

Recent studies \citep{dabah2024calibration, xi2024does} have investigated the integration of confidence calibration and CP to improve the quality of the prediction sets. Confidence calibration aims to ensure that the predicted probabilities of a classifier reflect its true accuracy \citep{xi2024does}. By calibrating the classifier's probabilities, one can obtain more reliable uncertainty estimates, which can potentially lead to more efficient prediction sets. 
However, the results presented in \citet{dabah2024calibration, xi2024does} show that the efficiency of prediction sets degrades when confidence calibration is combined with CP. This raises the question of how to effectively integrate accurate uncertainty quantification with guaranteed coverage to obtain efficient prediction sets.

We propose a novel approach that applies entropy-based reweighting to conformal classification in order to improve the efficiency of prediction sets. This method leverages the uncertainty of the classifier to dynamically adjust the weights, leading to more efficient prediction sets.

We conduct extensive experiments on various datasets, including AG News, CARER, MNIST, and Fashion MNIST, to evaluate the effectiveness of our approach. By comparing our method with existing techniques, we demonstrate its superior performance in terms of prediction efficiency and accuracy. Our experimental results highlight the robustness and applicability of the proposed method across different types of data and classification tasks.

\section{Related Work}
Previous related works can be considered under two categories:

\noindent
\textbf{(1) Conformal Prediction for Calibrated Models}

Methods like Platt's scaling \citep{platt1999probabilistic}, isotonic regression \citep{niculescu2005predicting}, spline-based probability calibration \citep{lucena2018spline}, and temperature scaling \citep{guo2017calibration} are used for post ad hoc calibration to adjust classifier confidence levels. 
Local Temperature Scaling \citep{ding2021local} calibrates probabilities in multi-label segmentation by assigning a unique temperature to each pixel for location-specific calibration.
Temperature-conditional Generative Flow Networks \citep{kim2023learning} incorporate a distinct pathway in their structure to adjust the policy's logits with the inverse temperature $\beta$, minimizing disruption to the network's parameters.
They lack the coverage guarantee offered by CP.
Moreover, temperature scaling, which is the most favored technique in this context, requires a considerable amount of validation data that must be representative of the training data. On the other hand, CP merely requires that the validation data (i.e., the calibration data) and the test data be exchangeable.
However, integrating confidence calibration with CP fails to deliver the anticipated advantages. \citet{dabah2024calibration, xi2024does} empirically show that using post ad hoc calibration before applying CP increase the size of prediction size. 
The observation that overconfident models (e.g. low temperature in temperature scaling) produce small prediction sets but an exceedingly low temperature fails to achieve the desired coverage \cite{xi2024does} motivates a novel loss function which penalizes the under coverage of the prediction set. 
\citet{cha2023on} studied the impact of temperature scaling in Bayesian graph neural networks on the prediction set produced by CP for node classification task.
As noted in \citet{stutz2023conformal}, in complex tasks like the dermatology problem, expert disagreements result in a high-entropy conditional probability $P(Y|X)$ that deviates from a one-hot distribution, leading to a coverage gap when modeled using a majority-voting scheme.

\noindent
\textbf{(2) Uncertainty Measures for Deep Classifiers}

Various methods for quantifying uncertainty have been introduced for neural network-based classifiers. These  uncertainty measures prove valuable in scenarios such as selecting samples based on uncertainty for active learning \citep{nguyen2022measure} and assigning uncertainty evaluations to individual data points \citep{chlaily2023measures, hullermeier2021aleatoric}. 
Additionally, uncertainty measures related to the concept of aleatoric and epistemic uncertainty have been discussed in \citet{gruber2023sources}. \citet{rossellini2024integrating} distinguishes between these two forms of uncertainty for constructing prediction intervals for conformalized quantile regressors. \citet{zhu2008active, nguyen2022measure} have explored the balance between uncertainty and representativeness in the context of active learning. 
Defining $f_Y(X)$ as the classifier's predictive probability of categorizing the object with feature $X$ as label $Y$, we can summarize some of existing uncertainty measures as follows:
\begin{enumerate}
    \item \textit{Entropy:} $-\sum_{Y=1}^{K} f_Y(X) \log f_Y(X)$.
    \item \textit{Smallest margin:} $\argmax_{Y^*\in \{1, \dots, K\}} f_{Y^*}(X) - \argmax_{Y\in \{1, \dots, K\}\setminus\{Y^*\}} f_Y(X)$.
    \item \textit{Gini impurity:} $\sum_{Y=1}^{K} (1-f_Y(X)) f_Y(X)$.
    \item \textit{Threshold \score{} \citep{sadinle2019least}:} $1 - f_Y(X)$, which penalizes cases where $f$ does not predict the observed label $Y$ with high probability. 
    \item \textit{Adaptive \score{} \citep{romano2020classification}:} 
    This is computed by summing up the sorted softmax values in a descending sequence
    \begin{equation}\label{eq:APS}
        a(\mathbf{f}(X), Y) = \sum_{i=1}^{r(Y, \mathbf{f}(X))-1} f_{(i)}(X) + U f_{(r(Y, \mathbf{f}(X)))}(X),
    \end{equation}
    where $f_{(1)}(X) > f_{(2)}(X) > \cdots > f_{(K)}(X)$ represent the order statistics of $\mathbf{f}(X)$, $U \sim \textrm{Uniform}(0, 1)$ is independent of everything else, and $r(Y, \mathbf{f}(X))$ is $\mathbf{f}(X)$'s ranking of the label $Y$. 
    \item \textit{Regularized Adaptive \score{} \citep{angelopoulos2021uncertainty}:}
    To tackle the long-tailed distribution issue inherent in softmax probabilities, Regularized Adaptive Prediction Sets (RAPS) eliminate classes that are less likely by imposing a penalty on classes that surpass a predetermined threshold:
    \begin{equation}\label{eq:RAPS}
        a(\mathbf{f}(X), Y) = \sum_{i=1}^{r(Y, \mathbf{f}(X))-1} \left(f_{(i)}(X) + \lambda \mathbbm{1}( i > k_{\textrm{reg}}) \right) + U f_{(r(Y, \mathbf{f}(X)))}(X),
    \end{equation}
    where $\lambda > 0$ discourages sets larger than $k_{\textrm{reg}}$.
    \item \textit{Rank-based conformity score \citep{luo2024trustworthy}:} The score function of RANK is defined as:
    \begin{align*}
        a(\mathbf{f}(X), Y) = \frac{r(Y, \mathbf{f}(X))}{K},
    \end{align*}
    which assigns a score based on the rank of the estimated probability $\hat f_Y(X)$ among all the estimated probabilities for feature $X$. The rank is divided by $K$ so that the range of the score is from 0 to 1. The prediction set gives higher priority to labels with larger ranks.
    
\end{enumerate}

\section{Entropy Reweighted Conformal Classification}
We start by defining key terms related to classification with CP. 
Let the feature space be $\mathbb{R}^d$ and the label space be $\mathcal{Y} = \{1, \dots, K \}$. Denote $f$ an $K$-class classification model and $D = \{ (X_n, Y_n) \in {\mathbb R}^d \times {\cal Y} \}_{n \in \mathcal{I}}$, ${\cal Y}$ a collection of i.i.d. random variables.
Assume $X_{N+1}$ is a test object with its label $Y_{N+1}$ masked. The output of the classification model, denoted as $f(X_{N+1}) \in [0, 1]^K$, represents the predicted probabilities of the test object belonging to each of the $K$ classes. 
A conformal Prediction Set (PS) at $X_{N+1}$, is a subset of the label space, $C(X_{N+1}) \subseteq {\cal Y}$ that obeys
\begin{align}
\label{marginal ps}
\textrm{ Prob}(Y_{N+1} \in C(X_{N+1})) \geq 1 - \alpha,
\end{align}
where $\alpha \in (0, 1)$ is a predefined confidence level and the probability is over $D$. 
We partition the set of indices $\mathcal{I}$ into $\mathcal{I}_1$ and $\mathcal{I}_2$, and then we build the PS using the split conformal method. 
The classification model $f$ is trained using the training samples $\{ (X_n, Y_n)\}_{n \in \mathcal{I}_1}$. The PS depends on the calibration samples $\{ (X_n, Y_n)\}_{n \in \mathcal{I}_2}$, the corresponding predicted probabilities, $\mathbf{f(X_n)} \in [0, 1]^K$, and an arbitrary conformity function, $a(\mathbf{f(X_n)}, Y_n)$ which evaluates the goodness of a model prediction compared with the corresponding label.
Marginal validity \eqref{marginal ps} holds if the test data and the calibration set, i.e., $(X_{N+1}, Y_{N+1})$ and $\{(X_n, Y_n)\}_{n \in \mathcal{I}_2}$ are exchangeable.

For classification problems, one can define a stronger property known as conditional coverage:
\begin{align}
\label{label conditional}
\textrm{ Prob}(Y_{N+1} \in C | Y_{N+1}=y) \geq 1 - \alpha,
\end{align}
which is alternatively termed as label-conditional coverage \citep{ding2024class, lofstrom2015bias} to differentiate it from the feature-conditional coverage \citep{einbinder2022training}:
\begin{align}
\label{feature conditional}
\textrm{ Prob}(Y_{N+1} \in C | X_{N+1}=x) \geq 1 - \alpha,
\end{align}

\subsection{Conformal Classification}
By definition, the PS based on conformity function $a(\mathbf{f(X_n)}, Y_n)$ is:
\begin{align}
C_A &= \{ y_{N+1} \in {\cal Y}, \sum_{n\in \mathcal{I}_2} \textbf{1}\left(A_n \leq A_{N+1} \right)\leq n_\alpha \}\
= \left\{ y_{N+1}\in {\cal Y}, A_{N+1} \leq Q_A \right\}
\end{align}
where $A_n = a(\mathbf{f(X_n)}, Y_n)$, $A_{N+1} = a(\mathbf{f}(X_{N+1}), y_{N+1})$, obeys \eqref{marginal ps} if $Q_A$ is the $(1-\alpha)$-th sample quantile of ${A_1, \dots, A_N }$ and $n_\alpha = \lceil (1 - \alpha) (|\mathcal{I}_2| + 1)\rceil$.
The meaning of the obtained PS depends on the definition of $a$.
In regression tasks, it is natural to let $a = a(\hat Y, Y)$ be the distance between predicted and observed labels.
In the classification setup, the model output is a discrete probability distribution.
Finding a conformity function that produces useful PS in the classification setup is less straightforward.
A popular choice, APS score (\ref{eq:APS}), is
\begin{align}\label{eq: conformity score}
A_n = a(\mathbf{f(X_n)}, Y_n) = \sum_{i=1}^{r(Y_n, \mathbf{f}(X_n))-1} f_{(i)}(X_n) + U f_{(r(Y_n, \mathbf{f}(X_n)))}(X_n),
\end{align}
where $f_{(i)}(X_n)$ denotes the $i$-th largest element of the probability vector $\mathbf{f}(X_n)$, $r(Y_n, \mathbf{f}(X_n))$ is the rank of the true label $Y_n$ in the probability vector $\mathbf{f}(X_n)$.

Although we use the APS score throughout our derivation and experiments, the entropy reweighting method is applicable and orthogonal to other score functions such as THR \citep{sadinle2019least}, RAPS \citep{angelopoulos2021uncertainty}, and SAPS \citep{huang2024conformal}.

\begin{algorithm}[htbp]
    \SetAlgoLined
    \SetEndCharOfAlgoLine{}
    \KwIn{Labeled data $\mathcal{D} = \{(X_n, Y_n)\}_{n\in [N]}$, unlabeled data $\mathcal{U} = \{X_{n'}\}_{n' \in [N']}$, coverage probability $1 - \alpha$, and a range of predefined temperature values $\{T_{j}\}_{j\in [M]}$.}
    \KwOut{Prediction set $C_{\alpha}$ for unlabeld data $\mathcal{U}$.}
    
    Randomly split $[N] = \{1, 2, \ldots, N\}$ into three (disjoint) parts $\mathcal{I}_1, \mathcal{I}_2, \mathcal{I}_3$. 
Set $\mathcal{D}_1 = \{(X_n, Y_n): i\in\mathcal{I}_1\}$, $\mathcal{D}_2 = \{(X_n, Y_n): n\in \mathcal{I}_2\}$, and $\mathcal{D}_3 = \{(X_n, Y_n): n\in \mathcal{I}_3\}$.\;

Train a classification model $f$ on $\mathcal{D}_1$ and use $f$ to obtain the logits on $\mathcal{D}_2$ and $\mathcal{D}_3$:
\begin{equation}
    \mathbf{z}(X_n) = \mathbf{f}(X_n), n\in \mathcal{I}_2 \cup \mathcal{I}_3.
\end{equation}

\For{each temperature $T_j$}{
    Compute the reweighted probability vectors $\tilde{\mathbf{f}}(X_n)$ for each data point in $\mathcal{D}_2$ according to (\ref{eq: entropy reweighted probs}):
    \begin{equation}
        \tilde{f}_k(X_n) = \frac{\exp(\frac{z_k(X_n)}{H(X_n) \cdot T_j})}{\sum_{i=1}^{K} \exp(\frac{z_i(X_n)}{H(X_n) \cdot T_j})}, n \in \mathcal{I}_2.
    \end{equation}\;
    
    Compute the reweighted conformity scores $\tilde{A}_n$ for each data point in $\mathcal{D}_2$ according to (\ref{eq: reweighted conformity score}):
    \begin{equation}\label{eq: reweighted score}
        \tilde{A}_n = \sum_{i=1}^{r(Y_n, \tilde{\mathbf{f}}(X_n))-1} \tilde{f}_{(i)}(X_n) + U \tilde{f}_{(r(Y_n, \tilde{\mathbf{f}}(X_n)))}(X_n), n\in \mathcal{I}_2.
    \end{equation}\;
    
    Construct the prediction sets $C_{\alpha}^{T_j}(X_n)$ for each data point in $\mathcal{D}_3$ as:
    \begin{equation}\label{reweighted prediction set}
        C_{\alpha}^{T_j}(X_n) = \left\{ y \in \mathcal{Y} : \tilde{A}_n(y) \geq Q_{\tilde{A}}^{T_j} \right\}, n \in \mathcal{I}_3,
    \end{equation}
    where $Q_{\tilde{A}}^{T_j}$ is the $(1 - \alpha)$-th sample quantile of $\{\tilde{A}_n : (X_n, Y_n) \in \mathcal{D}_2\}$ and $\tilde{A}_n(y) = \sum_{i=1}^{r(y, \tilde{\mathbf{f}}(X_n))-1} \tilde{f}_{(i)}(X_n) + U \tilde{f}_{(r(y, \tilde{\mathbf{f}}(X_n)))}(X_n)$.\;
    
    Calculate the average size of the resulting prediction sets:
    \begin{equation}
        \mathrm{AvgSize}(T_j) = \frac{1}{|\mathcal{D}_3|} \sum_{n \in \mathcal{I}_3} |C_{\alpha}^{T_j}(X_n)|.
    \end{equation}\;
}

Select the optimal temperature $T^* = \argmin\limits_{T_j} \mathrm{AvgSize}(T_j)$.\;

Compute the reweighted conformity scores $\tilde{A}_n$ for each $(X_n, Y_n) \in \mathcal{D}_2 \cup \mathcal{D}_3$ using $T^*$.\;

Compute the final prediction sets $C^{T^*}_{\alpha}(X)$ for each $X \in \mathcal{U}$ as:
\begin{equation}
    C_{\alpha}^{T^*}(X) = \left\{ y \in \mathcal{Y} : \tilde{A}(y) \geq Q_{\tilde{A}}^{T^*} \right\},
\end{equation}
where $\tilde{A}(y)$ is the reweighted conformity score for $X$ and label $y$, and $Q_{\tilde{A}}^{T^*}$ is the $(1 - \alpha)$-th sample quantile of $\{\tilde{A}_n(Y_i) : n\in \mathcal{I}_2 \cup \mathcal{I}_3\}$.\;

\Return the prediction sets $C_{\alpha}^{T^*}(X)$ for each $X \in \mathcal{U}$. 
\caption{Entropy Reweighted Conformal Prediction with Calibrated Temperature}
\label{alg:entropy-reweighted-conformal}

\end{algorithm}

\subsection{Reweighted Conformity Scores}\label{subsec: reweighted score}
As $C_A$ depends on all data points unregarding their attribute or label, the validity of $C$, i.e. the coverage guarantees in \eqref{marginal ps}, is marginal over the attribute and label spaces.
This means the uncertainty of the model on predicting the class probabilities is assumed to be constant.
As for the regression case, data heteroskedasticity may make such marginal PS highly inefficient.
In the Error Reweighted (ER) Conformal Prediction algorithm of  \citet{papadopoulos2008normalized}) and the localized conformal prediction of \citet{guan2023localized}, the regression conformity scores are rescaled by a pre-trained function to boost the adaptivity of the obtained prediction intervals.
More formally, ER computes the prediction intervals using a \emph{transformed conformity score}, $B_n = \frac{A_n}{\gamma + g(X_n)}$, where $g$ is a pre-trained model of the conditional residuals, i.e. $g(X_n) \approx \textrm{ E}_{Y_n|X_n}(|\mathbf{f(X_n)} - Y_n|)$ and $\gamma>0$ a regularization parameter.
The corresponding PS
\begin{align}\label{prediction set}
C_{B} = \{ y_{N+1}, B_{N+1} \leq Q_{B} \},
\end{align}
where $Q_{B}$ is the $(1-\alpha)$-th sample quantile of ${B_{\ntrain+1}, \dots, B_N }$.
When $A_n=|\mathbf{f(X_n)}-Y_n|$, this is equivalent to $B_n = \frac{A_n}{\gamma + \textrm{ E}_{A_n|X_n}(A_n)}$.
This work is about transferring the ER idea to the classification domain.
\citet{colombo2023training} considers a set of parameterized transformations of the conformity scores,
\begin{align}
B(X, Y) = b(A(X, Y), g(X, Y), \theta),
\end{align}
and propose to train them in a CP-aware sense.
In this formalism, the ER approach by \citet{papadopoulos2008normalized} corresponds to setting $b(a, g) = \frac{a}{\gamma + g}$. The benefit of ER approach is that it improves feature-conditional coverage (\ref{feature conditional}).

\subsection{Entropy-based Reweighting}
In the previous section, we emphasized the concept of reweighting the \score{s} based on the conditional residuals. However, since these residuals are estimated using a separate model $g$, there is a risk of model mis-specification if the model doesn't capture the underlying data distribution correctly. To address this, a logical approach would be to use information derived from the same model used for classification. 

A natural extension of the ER approach to the classification setting involves setting $g(X) = H(X)$, where the entropy inherently adapts to and reflects the classification model's uncertainty about an individual data point.

Denote $\mathbf{z(X)} = (z_1(X), \ldots, z_K(X))$ as the logit vector produced by the classification model, then the probability vector can be written as
\begin{equation}
    \mathbf{f}(X) = \frac{\mathbf{\textrm{exp}}(\mathbf{z(X)})}{\sum_{k=1}^{K} \textrm{exp} z_k(X)}.
\end{equation}

We consider reweighting the logit vector by the entropy of the corresponding probability vector. 
The entropy $H(X)$ of the probability vector $\mathbf{f}(X) = (f_1(X), \ldots, f_K(X))$ is defined as:
\begin{equation}
H(X) = -\sum_{k=1}^{K} f_k(X) \log f_k(X).
\end{equation}
Substituting the softmax function into the entropy formula, we have:
\begin{align}
H(X) &= -\sum_{k=1}^{K} \frac{\exp(z_k(X))}{\sum_{j=1}^{K} \exp(z_j(X))} \log \left(\frac{\exp(z_k(X))}{\sum_{j=1}^{K} \exp(z_j(X))}\right) \\
&= -\sum_{k=1}^{K} \frac{\exp(z_k(X))}{\sum_{j=1}^{K} \exp(z_j(X))} \left(z_k(X) - \log\left(\sum_{j=1}^{K} \exp(z_j(X))\right)\right) \\
&= -\sum_{k=1}^{K} f_k(X) z_k(X) + \log\left(\sum_{j=1}^{K} \exp(z_j(X))\right) \sum_{k=1}^{K} f_k(X) \\
&= -\sum_{k=1}^{K} f_k(X) z_k(X) + \log\left(\sum_{j=1}^{K} \exp(z_j(X))\right).
\end{align}

Adding another tunable temperature parameter $T$, the reweighted logit vector becomes:
\begin{equation}
\tilde{z}_k(X) = \frac{z_k(X)}{H(X)\cdot T}, \; k=1, \dots, K.
\end{equation}
The resulting reweighted probability vector can be obtained by applying the softmax function to the reweighted logits:
\begin{equation}\label{eq: entropy reweighted probs}
\tilde{f}_k(X) = \frac{\exp(\tilde{z}_k(X))}{\sum_{j=1}^{K} \exp(\tilde{z}_j(X))} = \frac{\exp(\frac{z_k(X)}{H(X)\cdot T})}{\sum_{j=1}^{K} \exp(\frac{z_j(X)}{H(X)\cdot T})}.
\end{equation}
The temperature parameter $T$ controls the sharpness of the reweighted probability distribution. When $T \rightarrow 0$, the distribution becomes more concentrated on the class with the highest reweighted logit. Conversely, when $T \rightarrow \infty$, the distribution becomes more uniform. By adjusting $T$, we can control the influence of entropy-based reweighting on the resulting probability distribution. In our experiments, we follow the cross-validation procedure outlined in \citet{yang2024selection} and the weighted aggregation idea in \citet{luo2024weightedaggregationconformityscores} to find the optimal temperature parameter using a separate validation set (see Algorithm \ref{alg:entropy-reweighted-conformal}).

The APS score (\ref{eq: conformity score}) for the entropy reweighted probability vector (\ref{eq: entropy reweighted probs}) thus becomes:
\begin{align}\label{eq: reweighted conformity score}
\tilde{A}_n = a(\tilde{\mathbf{f}}(X_n), Y_n) = \sum_{i=1}^{r(Y_n, \tilde{\mathbf{f}}(X_n))-1} \tilde{f}_{(i)}(X_n) + U \tilde{f}_{(r(Y_n, \tilde{\mathbf{f}}(X_n)))}(X_n),
\end{align}
where
$\tilde{\mathbf{f}}(X_n) = (\tilde{f}_1(X_n), \ldots, \tilde{f}_K(X_n))$ is the reweighted probability vector for input $X_n$ and $r(Y_n, \tilde{\mathbf{f}}(X_n))$ is the rank of the true label $Y_n$ in the reweighted probability vector $\tilde{\mathbf{f}}(X_n)$.
We present the full procedure for constructing prediction sets with the entropy reweighted method in Algorithm \ref{alg:entropy-reweighted-conformal}.

\section{Experiments}
In this section, we evaluate the performance of our proposed method. We conducted experiments on four datasets: AG News \citep{zhang2015characterlevel}, CelebA Attributes (CARER) \citep{liu2015faceattributes}, Fashion MNIST \citep{fmnist}, and MNIST \citep{lecun2010mnist}. For the AG News and CARER datasets, we use BERT as the classifier, while for the MNIST and Fashion MNIST datasets, we use a multi-layer neural network.

The experiment evaluates the calibration performance of classifiers trained on the four datasets. The calibration performance is assessed using various score functions, including the Entropy Reweighted (ER) score function proposed in this work, APS \citep{romano2020classification}, RAPS \citep{angelopoulos2021uncertainty}, and SAPS \citep{huang2024conformal}, across different desired coverage levels $1-\alpha$ ranging from 0.90 to 0.99 with a step size of 0.01. We employ the split conformal method for all score functions.
To account for variability, the experiment is repeated 10 times using different random splits of the data into calibration and test sets. The evaluation metrics used are coverage and size. Coverage measures the proportion of test instances for which the true label falls within the predicted prediction set, while size represents the average number of labels included in the prediction sets.

\begin{figure}[!htb]\label{fig:results}
    \centering
    \subfigure
    {%
        \includegraphics[width=0.45\textwidth]{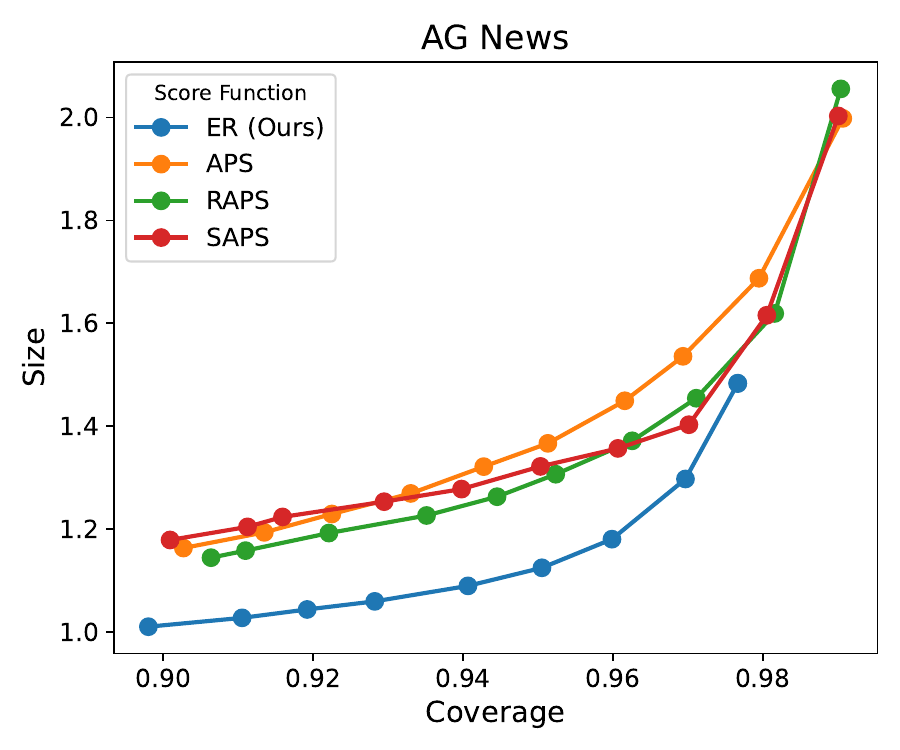}%
    }
    \hfill
    \subfigure
    {%
        \includegraphics[width=0.45\textwidth]{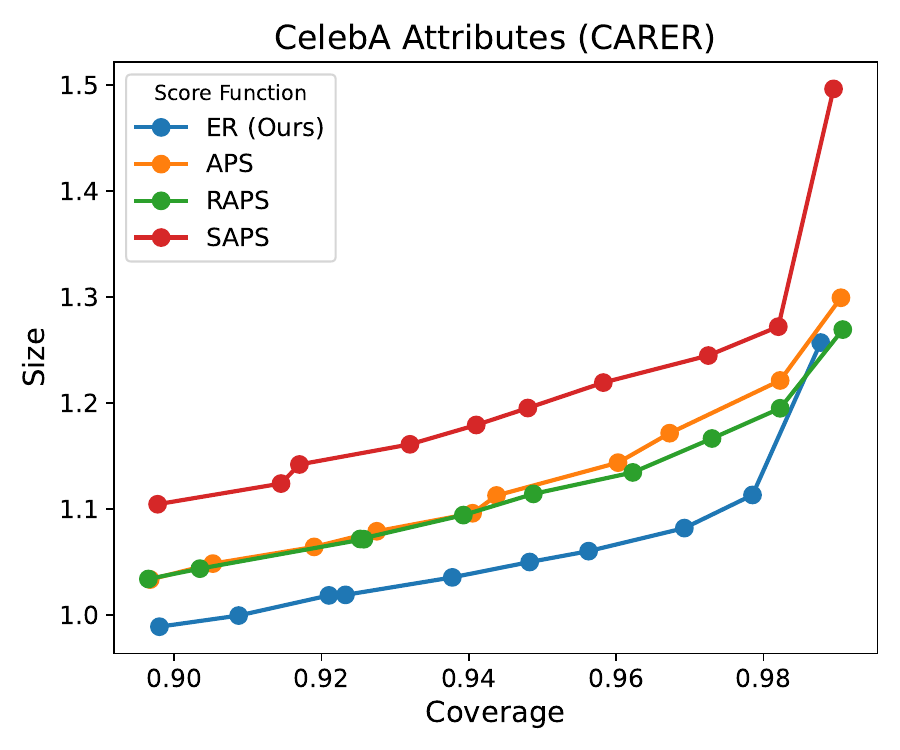}%
    }
    
    \vspace{1em}
    
    \subfigure
    {%
        \includegraphics[width=0.45\textwidth]{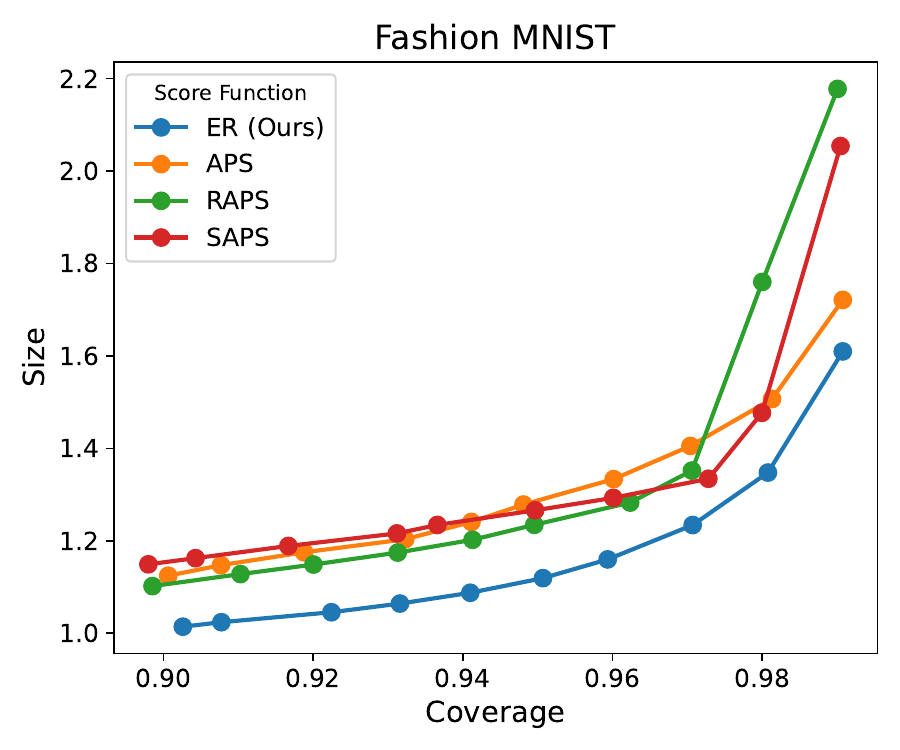}%
    }
    \hfill
    \subfigure
    {%
        \includegraphics[width=0.45\textwidth]{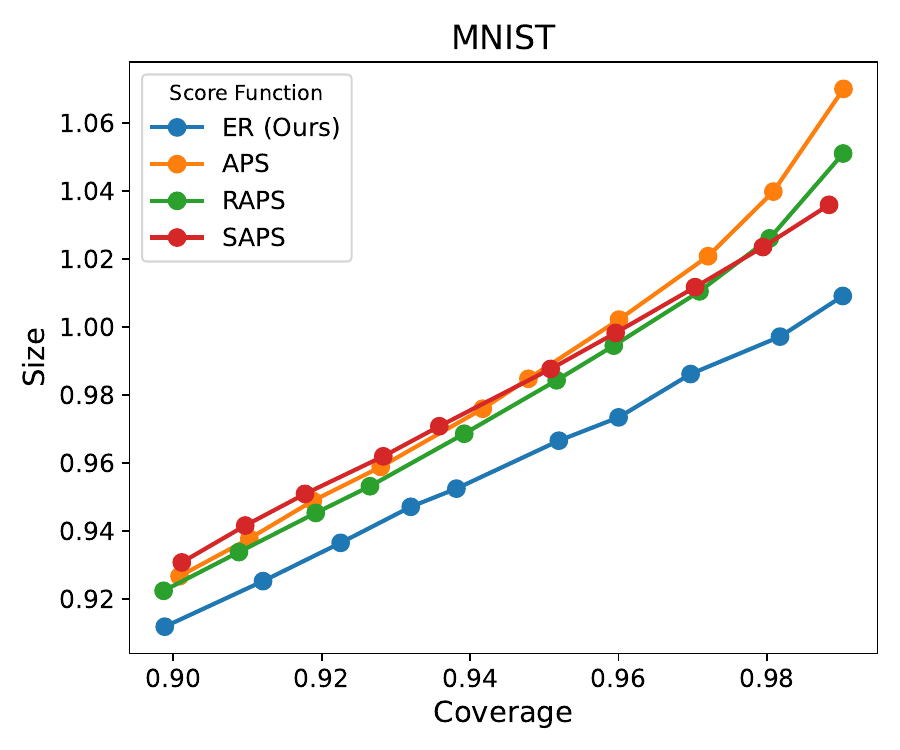}%
    }
    
    \caption{Size vs. Coverage plots for different datasets and score functions.}
    \label{fig:size_vs_coverage}
\end{figure}

The results are presented in Figure \ref{fig:results}, which shows the coverage-size plots for each dataset separately. Each plot displays the trade-off between coverage and size for different score functions. The proposed ER score function is labeled as ``ER (Ours)" in the plots. 
From the plots, we observe that the ER score function achieves competitive performance compared to other score functions across all datasets. It maintains good coverage while yielding relatively small confidence set sizes. The results demonstrate the effectiveness of the proposed entropy-based reweighting approach in improving the calibration performance of the classifiers.

Table 1 indicates that the proposed reweighting method outperforms the baseline model in terms of size across different $\alpha$ values.

To explain the superior performance of the proposed ER method, we categorize the cases into four scenarios: (1) the model was correct ($Y_n = \argmax_y f_y(X_n)$) and the entropy is low (indicating the model is certain about its prediction); (2) the model was correct and the entropy is high; (3) the model was incorrect and the entropy is low; and (4) the model was incorrect and the entropy is high. According to entropy reweighting, in the third scenario, the conformity score will adjust in preferable directions, meaning the score will increase and make the incorrect label harder to be included in the prediction set. Since this scenario is very common in over-confident deep network models, entropy reweighting will positively influence the scoring process and the construction of the prediction set.

\begin{table}[htb]
\resizebox{\textwidth}{!}{
    \centering
    \begin{tabular}{|l|l|c|c|c|c|c|c|}
        \hline
        \multirow{2}{*}{Dataset} & \multirow{2}{*}{Score Function} & \multicolumn{2}{c|}{$\alpha = 0.01$} & \multicolumn{2}{c|}{$\alpha = 0.05$} & \multicolumn{2}{c|}{$\alpha = 0.10$} \\
        \cline{3-8}
        & & Coverage & Size & Coverage & Size & Coverage & Size \\
        \hline
        \multirow{4}{*}{AG News} & ER (Ours) & 0.977 & \textbf{1.484} & 0.951 & \textbf{1.125} & 0.898 & \textbf{1.011} \\
        & APS & 0.991 & 1.998 & 0.951 & 1.367 & 0.903 & 1.163 \\
        & RAPS & 0.990 & 2.055 & 0.952 & 1.307 & 0.906 & 1.145 \\
        & SAPS & 0.990 & 2.003 & 0.950 & 1.322 & 0.901 & 1.179 \\
        \hline
        \multirow{4}{*}{CARER} & ER (Ours) & 0.988 & \textbf{1.257} & 0.948 & \textbf{1.050} & 0.898 & \textbf{0.989} \\
        & APS & 0.991 & 1.300 & 0.944 & 1.113 & 0.897 & 1.034 \\
        & RAPS & 0.991 & 1.270 & 0.949 & 1.115 & 0.897 & 1.034 \\
        & SAPS & 0.990 & 1.497 & 0.948 & 1.196 & 0.898 & 1.105 \\
        \hline
        \multirow{4}{*}{Fashion MNIST} & ER (Ours) & 0.991 & \textbf{1.610} & 0.951 & \textbf{1.119} & 0.903 & \textbf{1.014} \\
        & APS & 0.991 & 1.721 & 0.948 & 1.279 & 0.901 & 1.125 \\
        & RAPS & 0.990 & 2.178 & 0.950 & 1.235 & 0.899 & 1.102 \\
        & SAPS & 0.990 & 2.054 & 0.950 & 1.266 & 0.898 & 1.149 \\
        \hline
        \multirow{4}{*}{MNIST} & ER (Ours) & 0.990 & \textbf{1.009} & 0.952 & \textbf{0.967} & 0.899 & \textbf{0.912} \\
        & APS & 0.990 & 1.070 & 0.948 & 0.985 & 0.901 & 0.927 \\
        & RAPS & 0.990 & 1.051 & 0.952 & 0.984 & 0.899 & 0.923 \\
        & SAPS & 0.988 & 1.036 & 0.951 & 0.988 & 0.901 & 0.931 \\
        \hline
    \end{tabular}}
    \caption{Results for different $\alpha$ values.}
    \label{tab:results}
\end{table}

\section{Limitations and Future Work}
Although the experimental results indicate enhancements in both the conditional coverage and the efficiency of the PS, the current analysis remains largely empirical. We aim to enhance the theoretical robustness of the entropy reweighting approach by pursuing the following future directions.

\begin{enumerate}
    \item Temperature spline-based calibration: Instead of estimating a single temperature parameter $T$, we propose a temperature spline by estimating a feature-conditional temperature function $T(X)$. This added flexibility could enable us to enhance the conditional coverage further. 
    \item Sample reweighting approach: Furthermore, we consider to reweight the samples according to localized CP approach \citep{guan2023localized}, where we examine a local region around the test sample. Another option is by extending the calibration training idea in \citet{colombo2024normalizing} to classification settings.
    \item Application in graph-structure datasets: We plan to apply entropy reweighted method to graph-structured datasets to explore how graph topology influences model uncertainty quantification. We will focus on key applications such as node classification \citep{cha2023on}, link prediction \citep{luo2023anomalous}, and edge weight prediction \citep{luo2024conformal}, which is crucial for weighted graphs in various domains. 
\end{enumerate}

\end{document}